\documentclass[11pt]{article}

\usepackage[final]{acl}

\usepackage{times}
\usepackage{latexsym}
\usepackage[T1]{fontenc}
\usepackage[utf8]{inputenc}
\usepackage{microtype}
\usepackage{inconsolata}
\usepackage{graphicx}
\usepackage{url}

\title{mcdok at SemEval-2026 Task 13: Finetuning LLMs for Detection of Machine-Generated Code}

\author{Adam Skurla$^{1,2}$, Dominik Macko$^2$, Jakub Simko$^2$\\
  $^{1}$ Faculty of Information Technology, Brno University of Technology, Brno, Czechia\\
  $^{2}$ Kempelen Institute of Intelligent Technologies, Bratislava, Slovakia\\
 \texttt{\{adam.skurla, dominik.macko, jakub.simko\}}@kinit.sk \\}

\begin{document}
\maketitle
\begin{abstract}
Multi-domain detection of the machine-generated code snippets in various programming languages is a challenging task. SemEval-2026 Task~13 copes with this challenge in various angles, as a binary detection problem as well as attribution of the source. Specifically, its subtasks also cover generator LLM family detection, as well as a hybrid code co-generated by humans and machines, or adversarially modified codes hiding its origin. Our submitted systems adjusted the existing mdok approach (focused on machine-generated text detection) to these specific kinds of problems by exploring various base models, more suitable for code understanding. The results indicate that the submitted systems are competitive in all three subtasks. However, the margins from the top-performing systems are significant, and thus further improvements are possible.
\end{abstract}

\section{Introduction}

The advancements in large language models (LLMs) generation capabilities, also in generating programming code, make increasingly difficult to differentiate between human-written and machine-generated code. The SemEval-2026 Task~13~\citep{orel-etal-2026-semeval} deals with this problem in more challenging multi-generator, multi-domain, and multi-programming-language settings.

It consists of three subtasks, where the subtask A represents a binary machine-generated code detection. The training set contains three languages (C++, Python, Java) using the Algorithmic domain. Evaluation settings contain seen and unseen domains and languages.

Subtask B represents multi-class authorship detection, where the goal is to identify LLM family of the generator (out of 10 families) or human-written class (i.e., multiclass classification of 11 classes). The evaluation settings contain the generators seen in training as well as the generators unseen in the training (belonging to the LLM family that has been seen in training).

Subtask C represents hybrid code detection, where the goal is to distinguish between 4 classes: human-written, machine-generated, hybrid (i.e., partially written or completed by LLM), and adversarial (i.e., generated to mimic humans).

\begin{figure}[!t]
    \centering
    \includegraphics[width=\linewidth, trim=0.8cm 0.6cm 0.6cm 0.8cm, clip]{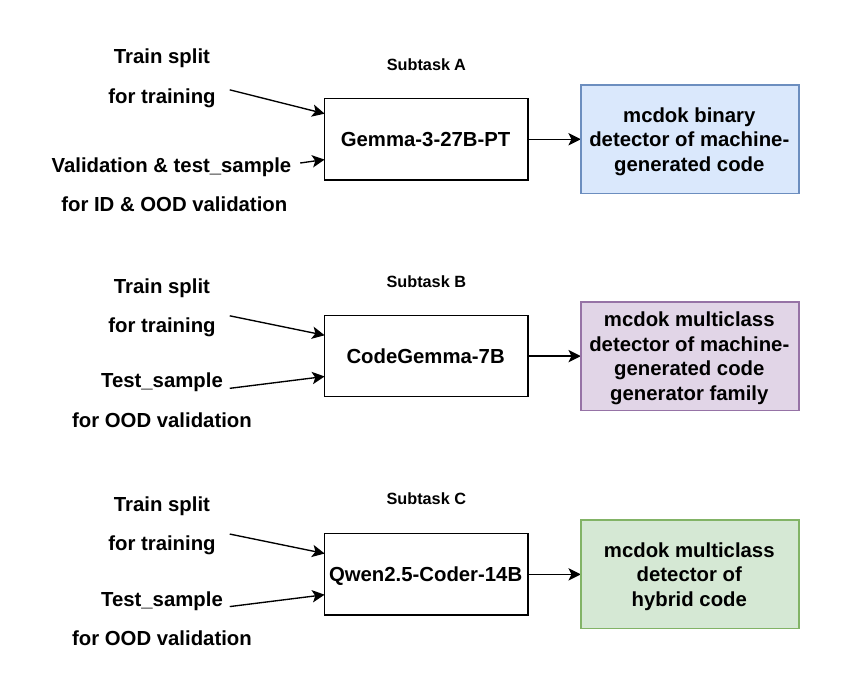}
    \caption{mcdok systems overview.}
    \label{fig:architecture}
\end{figure}

In all subtasks, our system originated in modification of the existing system mdok (machine detector of KInIT)~\citep{mdok}, which we call mcdok\footnote{\scriptsize\url{https://github.com/kinit-sk/mcdok-semeval2026}} (\textbf{m}achine-\textbf{c}ode \textbf{d}etector \textbf{o}f \textbf{K}InIT). The approach consisted in selection of more suitable base model for finetuning, which could better understand the programming code. The final submitted systems (see Figure~\ref{fig:architecture}) are based on Gemma-3-27B-PT~\citep{gemmateam2025gemma3technicalreport} for subtask A, CodeGemma-7B~\citep{codegemmateam2024codegemmaopencodemodels} for subtask B, and Qwen2.5-Coder-14B~\citep{hui2024qwen25codertechnicalreport} for subtask C.

\section{Background}

Our first experience with finetuning LLMs for a binary text-classification task was at SemEval-2024 Task 8~\citep{spiegel-macko-2024-kinit}, which was focused on detection of machine-generated text. We have further explored and increased the robustness of the used finetuning process~\citep{macko2025increasingrobustnessfinetunedmultilingual} using unique preprocessing steps and inclusion of obfuscated texts in the training. It eventually resulted into the mdok~\citep{mdok} finetuning approach, ranking 1st in both subtasks of PAN@CLEF2025~\citep{Bevendorff2025OverviewOT}, where
mdok was also extended to the multiclass scenario (within hybrid human-AI collaboration identification). It was further modified to the multilingual authorship attribution~\citep{lacava2025authorshipattributionmultilingualmachinegenerated}, utilizing traditional multi-class sequence classification finetuning approach. In this shared task, we are utilizing our experience in both, binary and multiclass classification, and apply them to a rather new domain of programming-code detection.

\section{System Overview}

As mentioned above, the mcdok system is heavily based on mdok~\citep{mdok}. Analogously to mdok, we have tried to keep the system as simple as possible -- i.e., to avoid ensembles.

The finetuning scripts have originated in the published mdok binary and multiclass versions\footnote{\url{https://github.com/kinit-sk/mdok}}. For binary detection, we have not used any text preprocessing (anonymization, lowercasing, homoglyphication) of the original mdok, since it would invalidate the code. We have reused innovative data selection, as described in Section~4. We have limited the training data to the official data provided by the organizers, as restricted by the shared task rules.

The finetuning process is based on the QLoRA~\citep{NEURIPS2023_1feb8787} parameter-efficient finetuning (PEFT) approach with 4-bit quantization using bitsandbytes and transformers\footnote{\url{https://github.com/huggingface/transformers}} python libraries. For LoRA, we have used the sequence classification PEFT task (i.e., no prompting), alpha of 16, dropout of 0.1, r of 64, and no bias. For finetuning, we have used a single sample per step (the batch size of 1 and no gradient accumulation), paged\_adamw\_32bit optimizer, learning rate of 2e-5 with the cosine scheduler type, a warmup ratio of 0.03, and validation each 100 steps (1000 steps for subtasks B and C). The finetuning process has taken up to 3 epochs with the final checkpoint selection based on the best metric achieved on validation set (Macro F1 for subtask A, loss for subtasks B and C). For calculation of the loss during finetuning, we have used weighted cross entropy loss, where weights have been set as inverse values of the class distribution in the training set.

We have published the source code (see footnote on the first page) for training as well as inference of the mcdok detectors, thus the models can be fully replicated.
The trained machine-code detectors can be easily applied by a user-friendly IMGTB framework~\citep{spiegel-macko-2024-imgtb}, just as other generic machine-text detectors.

\section{Experimental Setup}

Unless otherwise stated, the official train split of the data has been de-duplicated and subsampled to balance the classes and the ``test\_sample.parquet'' has been used for validation (i.e., checkpoint selection, as allowed by organizers in Kaggle discussion) during finetuning to represent the out-of-distribution (OOD) data (similarly do mdok approach).

In subtask A, the validation split combines the above mentioned OOD data with the official validation set. Afterwards, it has been subsampled up to 1k samples (if available) per each language and generator combination (to reduce the bias of majority), and out of such subset, it is further subsampled to balance the two classes (1k samples per class). Similarly for the train set, it is firstly subsampled up to 10k samples per language and generator and then up to 20k samples per class.

In subtask B, the validation split has been subsampled up to 1k samples per each generator (to have multiple generators belonging to the same class), and afterwards up to 500 samples for each class (smaller amount than in subtask A due to having 11 classes). For training it has been similarly subsampled firstly up to 10k samples per generator and afterwards up to 2k samples per class (a rather smaller number due to some classes being underrepresented).

In subtask C, the validation set has been subsampled up to 1k samples per each generator and language combination (analogously to subtask 1), and afterwards up to 500 samples per class. The train set has been subsampled firstly up to 10k samples per generator and language and then up to 10k samples per class.

The official evaluation metric for all subtasks is Macro F1 (macro average of F1 scores, representing harmonic means of precision and recall of each class).

\section{Results}

In subtask A, we have compared a bunch of base models to be finetuned for the task, ranging from 0.5B parameters size to 70B. The results on the official test set evaluated by the organizers are shown in Table~\ref{tab:performanceA}. Most of the examined models have been specialized for code understanding; however, interestingly, our best model, officially submitted to subtask A, is the general purpose Gemma-3-27B. However, by default (classification threshold of 0.5) it had rather random performance of 0.48 Macro F1. Based on our experience~\citep{spiegel-macko-2024-kinit}, we have also tried the detection based on fixing the classification threshold to probability of 1.0 (i.e., 100\% confidence in prediction of machine class), significantly boosting the performance for our finetuning approach. Analogously, this boost in performance (though in a lower scale) is observed for other three models (of 7B and 14B parameters), and in lower scale in the case of 70B model.

\begin{table}[!b]
\centering
\resizebox{0.8\linewidth}{!}{
\begin{tabular}{r|c}
\hline
\bfseries Detector & \bfseries Macro F1 \\
\hline
\textbf{gemma-3-27b-pt\_th1} & \textbf{0.69753} \\
codegemma-1.1-7b-it\_th1 & 0.64625 \\
codegemma-7b\_th1 & 0.64435 \\
Qwen2.5-Coder-14B\_th1 & 0.63818 \\
codegemma-7b & 0.59628 \\
Qwen2.5-Coder-7B & 0.57799 \\
CodeLlama-70b-hf\_th1 & 0.56363 \\
CodeLlama-70b-hf & 0.54776 \\
Qwen2.5-Coder-14B & 0.54498 \\
codegemma-1.1-7b-it & 0.54183 \\
Qwen2.5-Coder-32B & 0.51694 \\
Qwen2.5-Coder-0.5B & 0.49474 \\
gemma-3-27b-pt & 0.48006 \\
\hline
CodeBERT baseline & 0.30530 \\
random baseline & 0.50000 \\
\hline
\end{tabular}
}
\caption{The performance of the various system alternatives using the official test set for subtask A.}
\label{tab:performanceA}
\end{table}

\begin{figure}[!t]
    \centering
    \includegraphics[width=\linewidth] {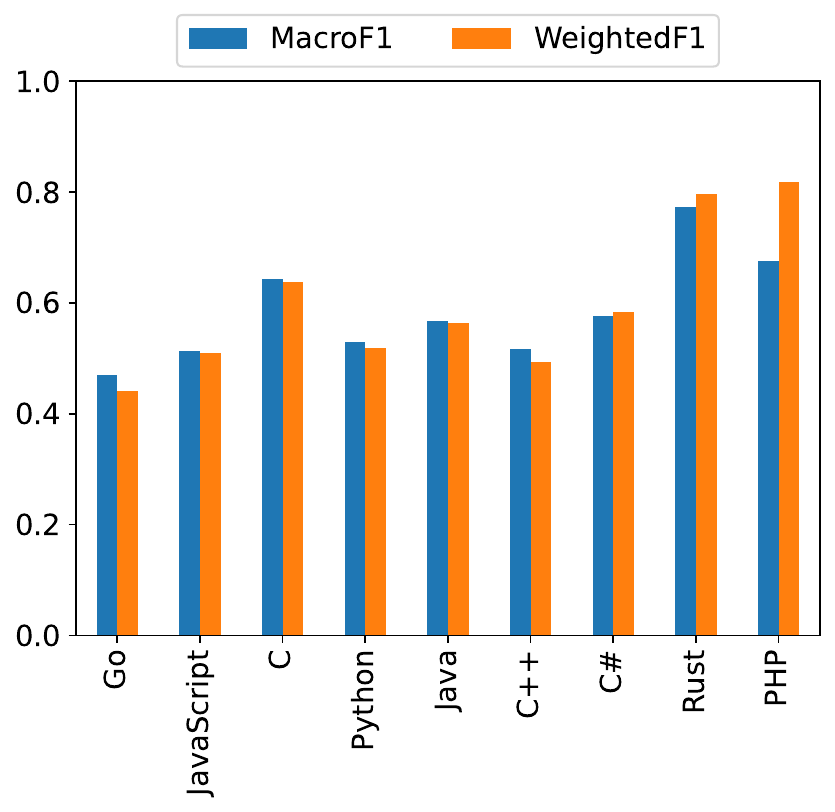} 
    \caption{Per-language performance of mcdok system in subtask A based on the filtered Droid test set.}
    \label{fig:analysisA}
\end{figure}
\begin{figure}[!t]
    \centering
    \includegraphics[width=\linewidth] {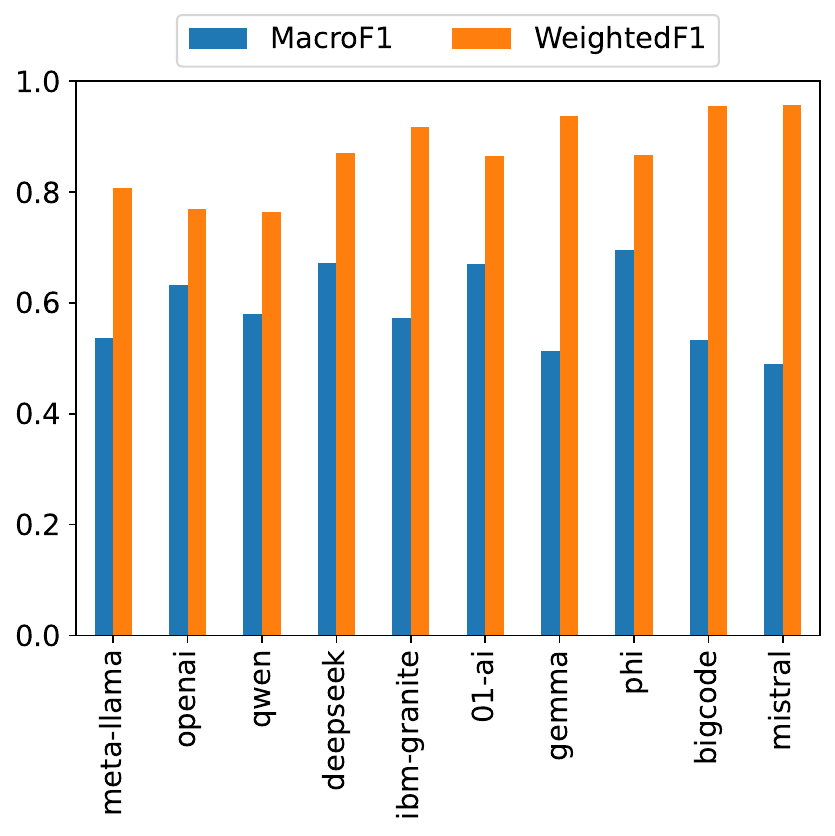} 
    \caption{Per-family performance of mcdok system in subtask A based on the filtered Droid test set.}
    \label{fig:analysisA_family}
\end{figure}

Since the organizers have not released the ground truth of the official test set (for analysis in the final system-description paper), we have analyzed the performance of the submitted system based on Droid test set~\citep{orel-etal-2025-droid}, on which the shared task is heavily based. To be sure to avoid data leakage and biased analysis, we have removed from such test set the samples that were included in the training or validation sets of the subtask A. The resulting filtered droid test set contains about 50k human as well as machine samples. The machine samples include also hybrid and adversarial samples. The results for performance comparison for individual (seen and unseen) programming languages are illustrated in Figure~\ref{fig:analysisA} and for individual LLM families in Figure~\ref{fig:analysisA_family}. Note that the data are not perfectly balanced, thus the comparison between dimensions might be biased (e.g., Rust, PHP, bigcode, and mistralai contain <1k samples of machine class). Besides Rust and PHP, a higher performance is observed for C and C\#, none of which has been included in the training. Higher differences between Weighted and Macro F1 scores in the comparison per LLM families indicate that the number of machine samples is much lower than the number of human samples (a high class imbalance).

For subtask B, the comparison of multiple system variants is provided in Table~\ref{tab:performanceB}. We experimented with several codegemma configurations that differed in model size and in the selection of training data. In particular, we compared training on the full dataset with preserved class distribution (codegemma models without the underscore character), undersampled balanced variants (``\_balanced''), and a setting that additionally incorporated the \textit{test\_sample} subset (``\_testsample'').

The models codegemma-7b and codegemma-2b were first trained on the full training dataset while keeping the original distribution of the 11 classes. The best result was achieved by codegemma-7b, with a Macro F1 of 0.396. The codegemma-2b model reached 0.365. The larger model achieved better performance in the multi-class generator attribution task.

We then trained balanced variants (codegemma-7b\_balanced and codegemma-2b\_balanced) using undersampling to reduce the number of samples in the dominant classes. This resulted in a more balanced class distribution and shorter training time. However, the balanced variants achieved lower Macro F1 scores than the models trained on the full dataset.


The submitted system in subtask B is based on codegemma of 7B parameters. This subtask of multi-class classification between 11 classes was difficult for our finetuning approach, but still the system outperformed both the official CodeBERT baseline as well as random baseline for 11 classes.

\begin{table}[!t]
\centering
\resizebox{0.85\linewidth}{!}{
\begin{tabular}{r|c}
\hline
\bfseries Detector & \bfseries Macro F1 \\
\hline
\textbf{codegemma-7b} & \textbf{0.39553} \\
codegemma-2b & 0.36475 \\
codegemma-7b\_balanced & 0.35714 \\
Qwen2.5-Coder-14B & 0.34415 \\
Qwen2.5-Coder-3B & 0.31463 \\
codegemma-2b\_balanced & 0.31188 \\
\hline
CodeBERT baseline & 0.22858 \\
random baseline & 0.09091 \\
\hline
\end{tabular}
}
\caption{The performance of the various system alternatives using the official test set for subtask B.}
\label{tab:performanceB}
\end{table}

\begin{figure}[!t]
    \centering
    \includegraphics[width=\linewidth] {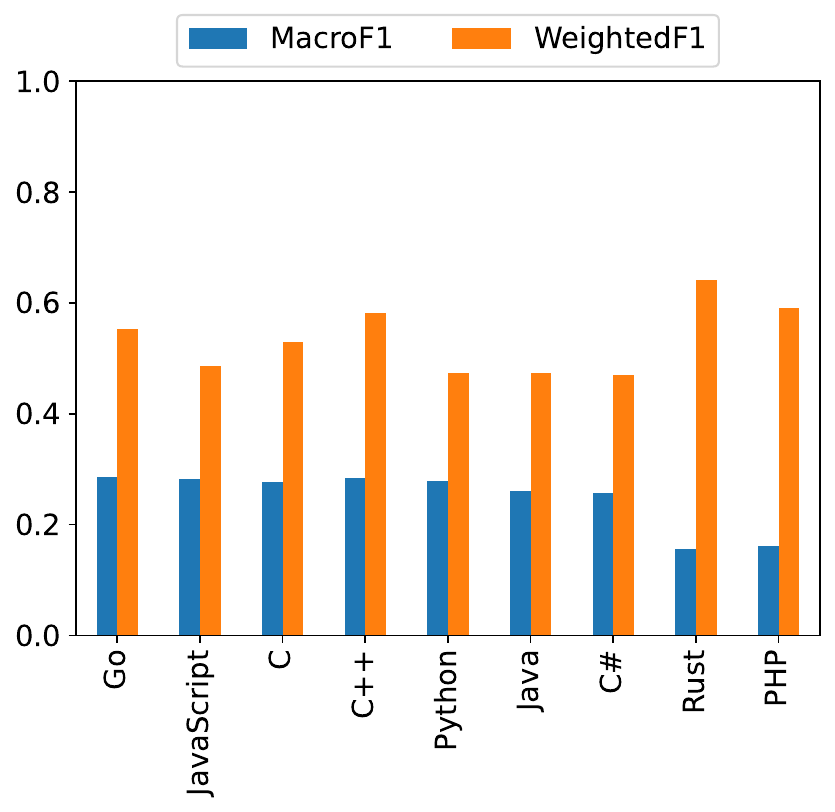} 
    \caption{Per-language performance (Macro F1) of mcdok system in subtask B based on the filtered Droid test set.}
    \label{fig:analysisB}
\end{figure}
\begin{figure}[!t]
    \centering
    \includegraphics[width=\linewidth] {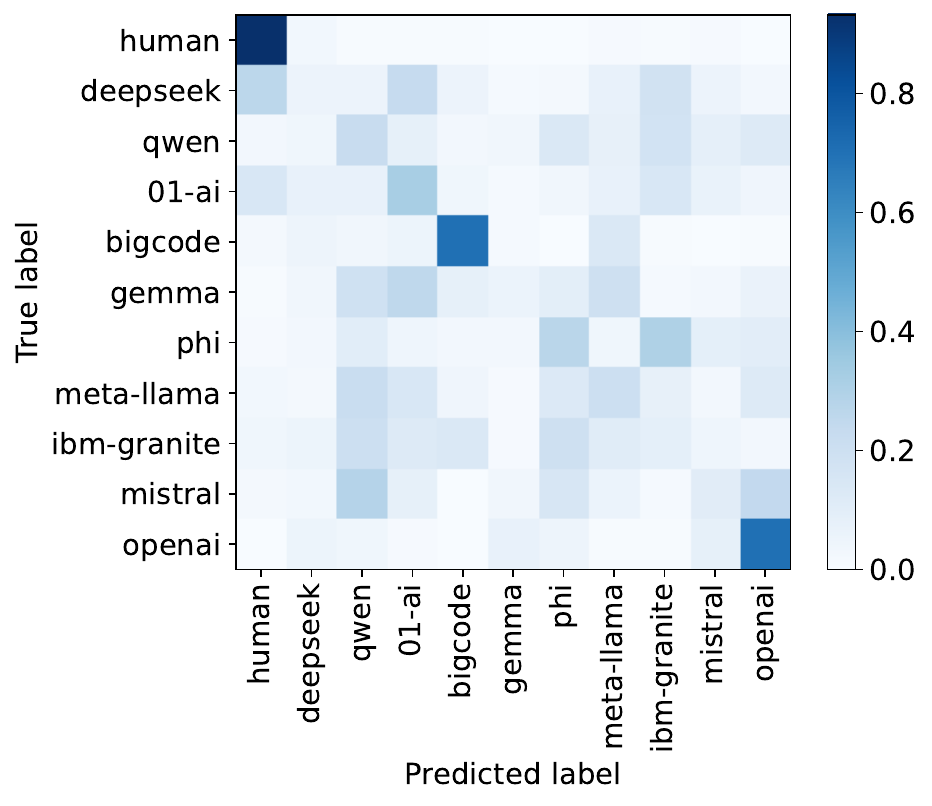} 
    \caption{Confusion matrix of mcdok system in subtask B based on the filtered Droid test set.}
    \label{fig:cmB}
\end{figure}

In subtask B, we have also analyzed the performance of the submitted system based on filtered (unseen) Droid test data, which for this subtask resulted in ~20k human samples, ~12k openai samples, ~10k qwen samples, bigcode and mistral included only <500 samples, while the other classes contained between 1k and 6k samples. The per-language results in Figure~\ref{fig:analysisB} indicate the stable performance across programming languages, with the Rust and PHP outliers due to missing samples for 4 classes (and <1k samples overall). The confusion matrix (Figure~\ref{fig:cmB}) indicates that the submitted system is strong in identifying human, bigcode, and openai classes, while it is confused in the others. A lower performance in comparison to the official test set is mainly because of the Droid test data also included adversarial and hybrid samples.

Finally, the comparison of four models for subtask C is provided in Table~\ref{tab:performanceC}. We have compared three variants of codegemma model and one of Qwen2.5-Coder. The last one represents the officially submitted system, performing the best on the official test set.

\begin{table}[!t]
\centering
\resizebox{0.7\linewidth}{!}{
\begin{tabular}{r|c}
\hline
\bfseries Detector & \bfseries Macro F1 \\
\hline
\textbf{Qwen2.5-Coder-14B} & \textbf{0.68643} \\
codegemma-1.1-7b-it & 0.65333 \\
codegemma-1.1-2b & 0.62094 \\
codegemma-7b & 0.59501 \\
\hline
CodeBERT baseline & 0.48120 \\
random baseline & 0.25000 \\
\hline
\end{tabular}
}
\caption{The performance of the various system alternatives using the official test set for subtask C.}
\label{tab:performanceC}
\end{table}

\begin{figure}[!t]
    \centering
    \includegraphics[width=\linewidth] {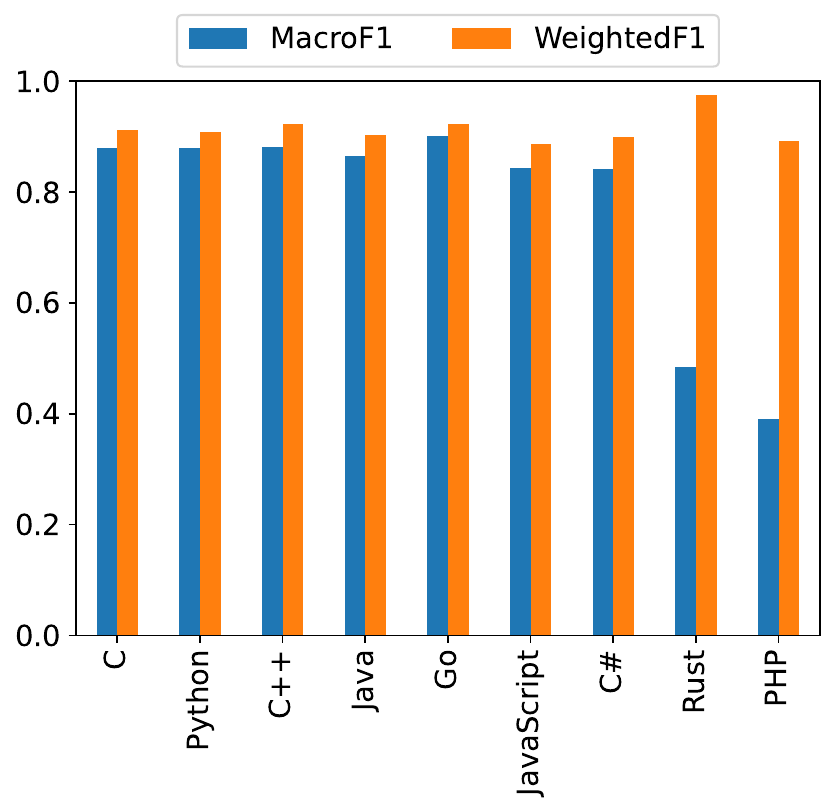} 
    \caption{Per-language performance (Macro F1) of mcdok system in subtask C based on the filtered Droid test set.}
    \label{fig:analysisC}
\end{figure}
\begin{figure}[!t]
    \centering
    \includegraphics[width=\linewidth] {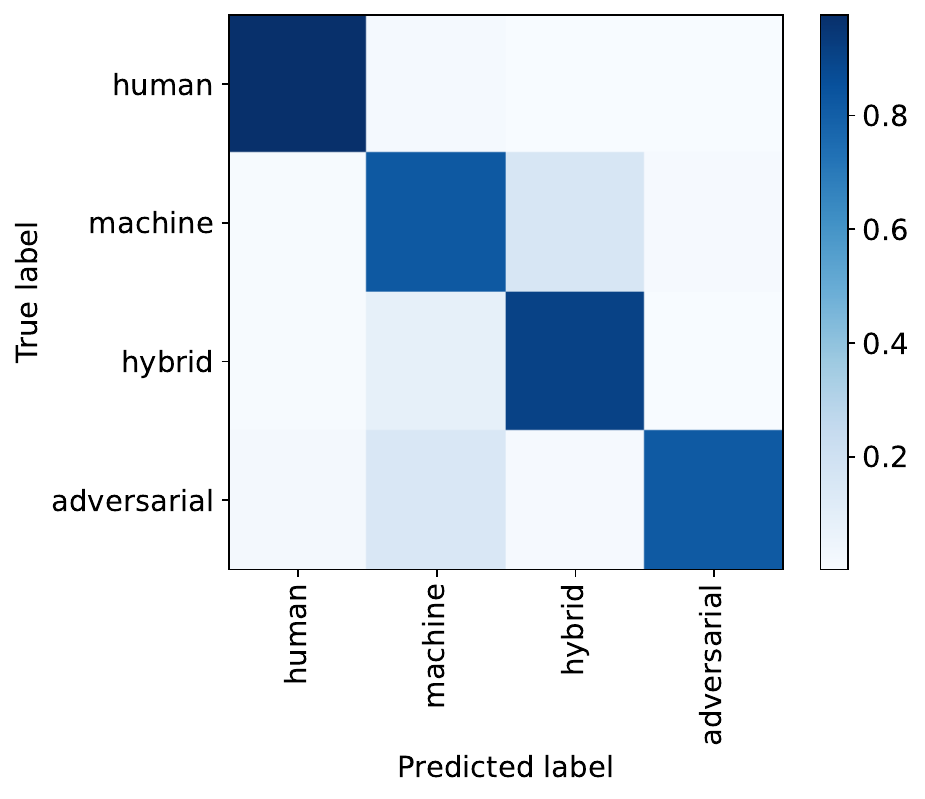} 
    \caption{Confusion matrix of mcdok system in subtask C based on the filtered Droid test set.}
    \label{fig:cmC}
\end{figure}

Similarly to previous, we have analyzed the performance of the submitted system based on filtered (unseen) Droid test data, including ~16k human samples, ~8k machine samples, ~4k hybrid samples, and ~5k adversarial samples. Rust and PHP have contained samples only for 2 of 4 classes, explaining the high difference in Weighted and Macro F1 scores (Figure~\ref{fig:analysisC}). Otherwise, the performance across programming languages has been quite stable. The system achieved the lowest F1 score for the hybrid class. The confusion matrix (Figure~\ref{fig:cmC}) indicates that the highest confusion was between machine and hybrid classes.

As shown above, the submitted systems outperformed the baselines (CodeBERT provided by organizers as well as random baseline based on number of classes). Based on the unofficial results (default kaggle leaderboards), the submitted systems ranked 10th of 81 submissions in subtask A (\textbf{88th percentile}), 13th of 34 submissions in subtask B (\textbf{62nd percentile}), and 5th of 32 submissions in subtask C (\textbf{84th percentile}).

\section{Conclusion}

Our work has shown, that generic machine-generated text detectors can be successfully transferred to machine-generated code detection task. We have prevalently focused on exploration of base models for finetuning. Interestingly, it resulted into selection of different models in each subtask as officially submitted systems. In general, codegemma-1.1-7b-it provides decent performance in all three subtasks. Further work might be focused on further care in sampling and balancing the train and validation sets, as well as on hyperparameters tuning of finetuning process.

\section*{Limitations}
Our experiments were focused mostly on code-specialized models; however, we have noticed a good performance of some general purpose models (in subtask A) as well. Therefore, the results reflect only the tested models and cannot be generalized. The other models might perform differently. The experiments were further limited by the official data and the used sampling techniques. Using different data might provide different results.

\section*{Acknowledgments}
Funded by the EU NextGenerationEU through the Recovery and Resilience Plan for Slovakia under the project No. 09I01-03-V04-00059.

\textbf{Computational resources}. This work was supported by the use of computational resources of the supercomputer PERUN, operated by the Supercomputing Centre at the Technical University of Košice (TUKE), Slovakia with the support of the European Union from the funds of the Recovery and Resilience Plan of the Slovak Republic within the framework of project No. 17I03-04-P03-00001, Development and design of a supercomputer for the National Supercomputing Center. We also acknowledge EuroHPC Joint Undertaking for awarding us access to Leonardo at CINECA, Italy. This work was also supported by the Ministry of Education, Youth and Sports of the Czech Republic through the e-INFRA CZ (ID:90254).

\bibliography{custom, anthology}

\appendix

\section{Computational Resources}
\label{sec:A}

For experiments regarding model fine-tuning and inference processes, we have used multiple GPU-accelerated systems. Experiments using $1\times$ NVIDIA A40 40GB GPU taken around 140 GPU hours, using $1\times$ NVIDIA A100 64GB GPU taken around 200 GPU hours, and using $1\times$ NVIDIA H200 140GB GPU taken around 800 GPU hours. Analysis has been done without the GPU acceleration.

\end{document}